\documentclass{ifac/ifacconf}

\usepackage{graphicx}      
\usepackage{natbib}        
\usepackage{comment}
\usepackage{amsfonts}
\usepackage{amsmath}
\usepackage{subcaption}
\usepackage{lipsum}
\usepackage{tikz}
\usepackage{float}
\usepackage{booktabs}
\usepackage{tcolorbox} 
\tikzset{
  traffic light/.pic={
    \draw[fill=black] (0,0) rectangle (0.4,1.2);
    \fill[green]   (0.2,1.0) circle (0.1);
    \fill[yellow](0.2,0.6) circle (0.1);
    \fill[red] (0.2,0.2) circle (0.1);
  }
}

\tikzset{
  pedestrian light/.pic={
    \draw[fill=black] (0,0) rectangle (0.3,0.35);
    \fill[red] (0.15,0.25) circle (0.07); 
    \draw[yellow, line width=0.5pt] (0.15,0.18) -- (0.15,0.08); 
    \draw[yellow, line width=0.5pt] (0.15,0.16) -- (0.10,0.10); 
    \draw[yellow, line width=0.5pt] (0.15,0.16) -- (0.20,0.10); 
    \draw[green, line width=0.5pt] (0.15,0.08) -- (0.10,0.02); 
    \draw[green, line width=0.5pt] (0.15,0.08) -- (0.20,0.02); 
  }
}

\begin{document}
\begin{frontmatter}

\title{Balancing Efficiency and Fairness in \\Traffic Light Control through \\Deep Reinforcement Learning\thanksref{footnoteinfo}} 

\thanks[footnoteinfo]{This work was partially carried out within the Italian National Center for Sustainable Mobility (MOST) and received funding from NextGenerationEU (Italian NRRP – CN00000023 - D.D. 1033 17/06/2022 - CUP C93C22002750006).}

\author{Matteo Cederle*,} 
\author{Giacomo Scatto*,} 
\author{Gian Antonio Susto}

\thanks[*]{M. Cederle and G. Scatto have equally contributed to this work.}

\address{University of Padova, 
   Italy \\ (e-mail: matteo.cederle@phd.unipd.it, giacomo.scatto@studenti.unipd.it, gianantonio.susto@unipd.it).}

\begin{abstract}                
Urban traffic congestion presents a significant challenge for modern cities, which impacts mobility and sustainability. Traditional traffic light control systems often fail to adapt to dynamic conditions, leading to inefficiencies. This paper proposes a novel deep reinforcement learning agent for traffic light control that addresses this limitation by explicitly integrating fairness considerations for both vehicular and pedestrian traffic. Unlike prior work, our approach dynamically balances these flows based on real-time demand, moving beyond systems focused solely on vehicles. Experimental results demonstrate that our agent effectively reduces congestion while ensuring equitable service for both the categories of road users. This research contributes to a practical and adaptable solution for intelligent traffic management within the framework of smart cities, paving the way for more efficient and inclusive urban mobility.
\end{abstract}


\end{frontmatter}

\section{Introduction}
\label{sec:intro}

Urbanization is an ever-growing trend in modern cities. The development of large urban centers undoubtedly generates advantages from an economical and technological perspective. However, on the other hand, it also introduces critical challenges from a social, organizational and ecological point of view. 
Road traffic stands among those challenges as an issue that should not be overlooked; urban centers face daily gridlocks, impacting not only individual mobility, but also public and commercial transportation. 

However, the dualistic perspective of the consequences brought by the technological advancement offers us also the instruments to address the traffic management problem: the application of technological innovations at a urban level gives raise to the concept of \textit{smart cities}, defined as geographical areas in which high technologies such as ICT, logistics, energy production, and so on, cooperate to create benefits for citizens (\cite{dameri2013searching}). 
Within this smart-city framework, traffic management plays a crucial role. To effectively reduce congestion,
it is necessary to fully understand traffic patterns and have a wide comprehensions of its dynamics. 


One of the key steps towards the improvement of traffic management focuses on the optimization of traffic light systems. Traditional traffic lights operate on fixed timers, often failing to account for real-time traffic conditions. This results in inefficient green light durations, unnecessary stops, and extended queues. The inefficiencies of these static systems become more evident during peak hours or unexpected surges in traffic volume (\cite{allsop1972delay}).
A first step towards the development of smart traffic control schemes is represented by actuated traffic lights, developed to overcome the limitation of fixed-time schemes. They exploit sensors, like inductive loops embedded in the pavement, to detect the presence of vehicles and adjust phases accordingly. 
Despite being more responsive than fixed-time systems, their decision-making capabilities remain local and still based on hand-crafted rules. 
As a result, they inevitably fail to coordinate traffic in a consistently effective way.

To address these issues, an important volume of research has been done in academic and industry communities to build adaptive traffic signal control systems. 
With the advancements in the field of reinforcement learning (RL), traffic control systems are starting to become smarter and more efficient, enabling real-time optimization of traffic flows and significantly reducing congestion 
(\cite{alegre2019sumo, li2016traffic, wei2018intellilight, mousavi2017traffic, liang2019deep, yazdani2023intelligent, aslani2017adaptive}). These works vary under several key aspects. Firstly, in the state representation, including hand-crafted features such as lane density, queue length, traffic light configuration, or image-like features as position and velocity. Secondly, they may differ in the reward design, including waiting time, average delay, queue length and outflow rate.
Finally, in terms of action space, the aforementioned works implement two main approaches: exploiting binary actions, i.e. maintain phase or switch to the next one, or extending the action space by indicating the time duration the current phase, based on a pool of options.

Despite significant progress in recent years, only a limited number of studies (\cite{yazdani2023intelligent, zhu2022context}) have examined the integration of vehicle and pedestrian flow within traffic-light control systems. This oversight underscores the need for the current investigation. The primary objective of our study is to introduce \textit{fairness} into the system, by enabling the model to dynamically assess and prioritize the two aforementioned flows based on real-time demand. In this context, fairness refers to avoiding systematic over-prioritization of one category of road users at the expense of another.
Recent developments in smart mobility have brought increasing attention to the role of fairness in intelligent transportation systems (\cite{cederle2025fairness, salazar2024accessibility}). Within the domain of traffic-light control, fairness has been explored mainly in relation to vehicle flows (\cite{li2020fairness, ye2022fairlight}), yet, to the best of our knowledge, only limited attention has been devoted to explicitly modeling fairness between pedestrian and vehicular flows.
We therefore propose a system designed to balance both types of traffic, depending on the specific circumstances. 

The main contributions brought by this study can be summarized as follows:
\begin{enumerate}
    \item We propose a novel RL agent for traffic light control, by combining state-of-the-art techniques, while requiring at the same time realistic observations, to ensure the real world applicability of our method.
    \item We explicitly consider the presence of pedestrians in the environment, by augmenting our algorithm with fairness considerations, to ensure equitable service for both the categories of road users. 
    \item We implement our approach in simulation, and we show its effectiveness in learning policies capable of reducing congestion while promoting at the same time fairness of service to both vehicles and pedestrians.
\end{enumerate}

The structure of this study is organized as follows: Section 2 provides an overview of essential concepts in deep reinforcement learning, while Section 3 elaborates on the main contributions of our research. In Section 4, we present the outcomes from various simulated experiments, which assess the effectiveness of our approach. Lastly, Section 5 concludes by summarizing our findings and suggesting potential avenues for future work.

\section{Background}
\label{sec:backgr}

Reinforcement learning is a branch of machine learning in which an \textit{agent} iteratively learns to solve a \textit{task} by interacting with its \textit{environment}. The goal is to maximize the \textit{cumulative rewards} obtained over time. A RL problem is formalized by the framework of the \textit{Markov Decision Process} (MDP, \cite{sutton2018reinforcement}), which defines the problem as a five-element tuple: ${\left\langle \mathcal{S}, \mathcal{A}, \mathcal{P}, \mathcal{R}, \gamma \right\rangle}$. Here, $\mathcal{S}$ and $\mathcal{A}$  represent the \textit{state} and \textit{action spaces}, respectively. The agent interacts with the environment through a policy $\pi:~\mathcal{S} \to \mathcal{A}$, which selects actions that influence future states. The \textit{state transition function} $\mathcal{P}$ models the system’s dynamics: 
$\mathcal{P}(s, a, s') = P\left[S_{t+1} = s' \mid S_t = s, A_t = a \right]$.
Each transition is evaluated through a \textit{reward function} $\mathcal{R}(s, a, s'): \mathcal{S} \times \mathcal{A} \times \mathcal{S} \to \mathbb{R}$, which provides information on the quality of the chosen action. Additionally, the \textit{discount factor} $\gamma \in [0,1)$ controls the importance of future rewards, shaping the \textit{cumulative return} $G_t=\sum_{k=0}^\infty \gamma^kr_{t+k+1}$.
Solving a RL problem coincides with finding the optimal policy that maximizes the action-value function $Q_\pi(s,a)=\mathbb{E}_\pi[G_t|S_t=s,A_t=a]$. In high-dimensional or complex environments, the action-value function is often parameterized by a neural network with parameters $\theta$, i.e. $Q\to Q_\theta$. To break the correlations between consecutive samples, a \textit{replay memory} is used to store the transition tuples $(s,a,r,s')$ at each time step. Finally, the action-value network $Q_\theta$ is optimized through the loss in Equation~(\ref{eq:loss}), inspired by the temporal difference error of Q-learning (\cite{sutton2018reinforcement}):
\begin{equation}
\label{eq:loss}
    \mathcal{L}(\theta) = \cfrac{1}{|\mathcal{B}|}\sum_{i\in \mathcal{B}} [(r_i+\gamma \max_{a'}Q(s_i',a';\bar{\theta}) - Q(s_i,a_i;\theta))^2].
\end{equation}
Here, $\mathcal{B}$ is a batch of transitions sampled from the replay memory and $\bar{\theta}$ denotes the parameters of a target network, which are slowly updated to stabilize training. All these components originate the \textit{Deep Q-Networks (DQN)} algorithm, introduced in \cite{mnih2015human}. As a final remark, DQN is nowadays usually implemented considering the \textit{double network} modification (DDQN), useful to reduce the overestimation bias problem (\cite{van2016deep}). We refer to the original paper for a detailed explanation of its implementation.

\begin{figure}[h]
\begin{center}
\begin{tikzpicture}
    \fill[gray] (-4,-2) rectangle (4,2); 
    \fill[gray] (-2,-4) rectangle (2,4); 
    
    \draw[white, dashed, line width=2pt] (0.6,-4) -- (0.6,-1.8);
    \draw[white, dashed, line width=2pt] (1.2,-4) -- (1.2,-1.8);
    \draw[white, dashed, line width=2pt] (-0.6,-4) -- (-0.6,-1.8);
    \draw[white, dashed, line width=2pt] (-1.2,-4) -- (-1.2,-1.8);

    \draw[white, dashed, line width=2pt] (0.6,4) -- (0.6,1.8);
    \draw[white, dashed, line width=2pt] (1.2,4) -- (1.2,1.8);
    \draw[white, dashed, line width=2pt] (-0.6,4) -- (-0.6,1.8);
    \draw[white, dashed, line width=2pt] (-1.2,4) -- (-1.2,1.8);

    \draw[white, line width=2pt] (-4,0) -- (-1.8,0);
    \draw[white, line width=2pt] (1.8,0) -- (4,0);
    \draw[white, line width=2pt] (0,-4) -- (0,-1.8);
    \draw[white, line width=2pt] (0,1.8) -- (0,4);

    \draw[white, dashed, line width=2pt] (-4,-1.2) -- (-1.8,-1.2);
    \draw[white, dashed, line width=2pt] (-4,-0.6) -- (-1.8,-0.6);
    \draw[white, dashed, line width=2pt] (-4,0.6) -- (-1.8,0.6);
    \draw[white, dashed, line width=2pt] (-4,1.2) -- (-1.8,1.2);

    \draw[white, dashed, line width=2pt] (1.8,-1.2) -- (4,-1.2);
    \draw[white, dashed, line width=2pt] (1.8,-0.6) -- (4,-0.6);
    \draw[white, dashed, line width=2pt] (1.8,0.6) -- (4,0.6);
    \draw[white, dashed, line width=2pt] (1.8,1.2) -- (4,1.2);
    
    \draw[white, line width=2pt] (-4,-1.8) -- (-1.8,-1.8);
    \draw[white, line width=2pt] (0,-1.8) -- (4,-1.8);
    \draw[white, line width=2pt] (-4,1.8) -- (0,1.8);
    \draw[white, line width=2pt] (1.8,1.8) -- (4,1.8);
    \draw[white, line width=2pt] (-1.8,-4) -- (-1.8,0);
    \draw[white, line width=2pt] (-1.8,1.8) -- (-1.8,4);
    \draw[white, line width=2pt] (1.8,-4) -- (1.8,-1.8);
    \draw[white, line width=2pt] (1.8,0) -- (1.8,4);

    \draw[white, line width=2pt] (1.4,-2.4) -- (1.4,-1.8);
    \draw[white, line width=2pt] (1.6,-2.4) -- (1.6,-1.8);
    \draw[white, line width=2pt] (1,-2.4) -- (1,-1.8);
    \draw[white, line width=2pt] (0.8,-2.4) -- (0.8,-1.8);
    \draw[white, line width=2pt] (0.6,-2.4) -- (0.6,-1.8);
    \draw[white, line width=2pt] (0.4,-2.4) -- (0.4,-1.8);
    \draw[white, line width=2pt] (0.2,-2.4) -- (0.2,-1.8);
    \draw[white, line width=2pt] (1.2,-2.4) -- (1.2,-1.8);

    \draw[white, line width=2pt] (-1.4,-2.4) -- (-1.4,-1.8);
    \draw[white, line width=2pt] (-1.6,-2.4) -- (-1.6,-1.8);
    \draw[white, line width=2pt] (-1,-2.4) -- (-1,-1.8);
    \draw[white, line width=2pt] (-0.8,-2.4) -- (-0.8,-1.8);
    \draw[white, line width=2pt] (-0.6,-2.4) -- (-0.6,-1.8);
    \draw[white, line width=2pt] (-0.4,-2.4) -- (-0.4,-1.8);
    \draw[white, line width=2pt] (-0.2,-2.4) -- (-0.2,-1.8);
    \draw[white, line width=2pt] (-1.2,-2.4) -- (-1.2,-1.8);

    \draw[white, line width=2pt] (-1.4,2.4) -- (-1.4,1.8);
    \draw[white, line width=2pt] (-1.6,2.4) -- (-1.6,1.8);
    \draw[white, line width=2pt] (-1,2.4) -- (-1,1.8);
    \draw[white, line width=2pt] (-0.8,2.4) -- (-0.8,1.8);
    \draw[white, line width=2pt] (-0.6,2.4) -- (-0.6,1.8);
    \draw[white, line width=2pt] (-0.4,2.4) -- (-0.4,1.8);
    \draw[white, line width=2pt] (-0.2,2.4) -- (-0.2,1.8);
    \draw[white, line width=2pt] (-1.2,2.4) -- (-1.2,1.8);

    \draw[white, line width=2pt] (1.4,2.4) -- (1.4,1.8);
    \draw[white, line width=2pt] (1.6,2.4) -- (1.6,1.8);
    \draw[white, line width=2pt] (1,2.4) -- (1,1.8);
    \draw[white, line width=2pt] (0.8,2.4) -- (0.8,1.8);
    \draw[white, line width=2pt] (0.6,2.4) -- (0.6,1.8);
    \draw[white, line width=2pt] (0.4,2.4) -- (0.4,1.8);
    \draw[white, line width=2pt] (0.2,2.4) -- (0.2,1.8);
    \draw[white, line width=2pt] (1.2,2.4) -- (1.2,1.8);

    \draw[white, line width=2pt] (-2.4,-1.6) -- (-1.8,-1.6);
    \draw[white, line width=2pt] (-2.4,-1.4) -- (-1.8,-1.4);
    \draw[white, line width=2pt] (-2.4,-1.2) -- (-1.8,-1.2);
    \draw[white, line width=2pt] (-2.4,-1) -- (-1.8,-1);
    \draw[white, line width=2pt] (-2.4,-0.8) -- (-1.8,-0.8);
    \draw[white, line width=2pt] (-2.4,-0.6) -- (-1.8,-0.6);
    \draw[white, line width=2pt] (-2.4,-0.4) -- (-1.8,-0.4);
    \draw[white, line width=2pt] (-2.4,-0.2) -- (-1.8,-0.2);

    \draw[white, line width=2pt] (-2.4,1.6) -- (-1.8,1.6);
    \draw[white, line width=2pt] (-2.4,1.4) -- (-1.8,1.4);
    \draw[white, line width=2pt] (-2.4,1.2) -- (-1.8,1.2);
    \draw[white, line width=2pt] (-2.4,1) -- (-1.8,1);
    \draw[white, line width=2pt] (-2.4,0.8) -- (-1.8,0.8);
    \draw[white, line width=2pt] (-2.4,0.6) -- (-1.8,0.6);
    \draw[white, line width=2pt] (-2.4,0.4) -- (-1.8,0.4);
    \draw[white, line width=2pt] (-2.4,0.2) -- (-1.8,0.2);

    \draw[white, line width=2pt] (2.4,1.6) -- (1.8,1.6);
    \draw[white, line width=2pt] (2.4,1.4) -- (1.8,1.4);
    \draw[white, line width=2pt] (2.4,1.2) -- (1.8,1.2);
    \draw[white, line width=2pt] (2.4,1) -- (1.8,1);
    \draw[white, line width=2pt] (2.4,0.8) -- (1.8,0.8);
    \draw[white, line width=2pt] (2.4,0.6) -- (1.8,0.6);
    \draw[white, line width=2pt] (2.4,0.4) -- (1.8,0.4);
    \draw[white, line width=2pt] (2.4,0.2) -- (1.8,0.2);

    \draw[white, line width=2pt] (2.4,-1.6) -- (1.8,-1.6);
    \draw[white, line width=2pt] (2.4,-1.4) -- (1.8,-1.4);
    \draw[white, line width=2pt] (2.4,-1.2) -- (1.8,-1.2);
    \draw[white, line width=2pt] (2.4,-1) -- (1.8,-1);
    \draw[white, line width=2pt] (2.4,-0.8) -- (1.8,-0.8);
    \draw[white, line width=2pt] (2.4,-0.6) -- (1.8,-0.6);
    \draw[white, line width=2pt] (2.4,-0.4) -- (1.8,-0.4);
    \draw[white, line width=2pt] (2.4,-0.2) -- (1.8,-0.2);

\pic[scale=0.8, rotate=90] at (-0.45,1.35) {traffic light};           
\pic[scale=0.8] at (1.35,0.45) {traffic light};            
\pic[scale=0.8, rotate=180] at (-1.35,-0.45) {traffic light}; 
\pic[scale=0.8, rotate=-90] at (0.45,-1.35) {traffic light};  

\pic[rotate=0] at (-2.3, -2.35) {pedestrian light};
\pic[rotate=0] at (2, -2.35) {pedestrian light};
\pic[rotate=0] at (-2.3, 2) {pedestrian light};
\pic[rotate=0] at (2, 2) {pedestrian light};

\end{tikzpicture}
\end{center}
\caption{Schematic representation of the four-way three-lane signalized intersection considered for this study.}
\label{fig:int}
\end{figure}
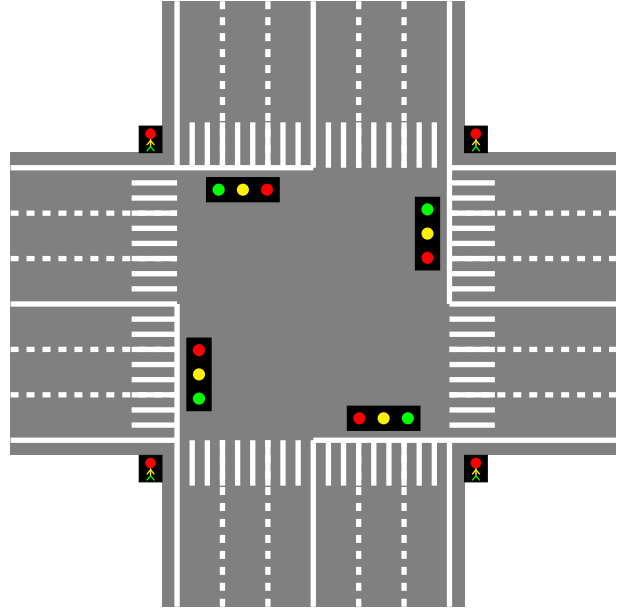

\section{Proposed Approach}
\label{sec:method}

The environment considered for this work consists of a signalized four-way three-lane intersection, as shown in Figure \ref{fig:int}. As already discussed in Section \ref{sec:intro}, it is populated by both vehicles and pedestrians, and our objective is to reduce congestion while ensuring at the same time fairness of service for both categories of road users. In the remainder of this section, we will describe the building blocks of the MDP formalization designed for~this work.

\subsection{State and action spaces representation}

Taking inspiration from \cite{alegre2019sumo}, and extending their intuitions to suit our specific objectives, we designed a state space composed by five components, which all together provide the RL agent with a comprehensive understanding of the current situation in the environment. The first two components are internal parameters to the intersection controller, i.e. the currently active \textit{traffic light phase} $a\in\mathcal{A}$, where $\mathcal{A}$ is the finite discrete set containing the available phases, and the \textit{elapsed time} since the last phase change $T\in\mathbb{R}^+$. Moreover, the state also includes two different vehicle-related information, i.e. the \textit{density} $d\in\mathbb{R}^{4\times3}_{[0,1]}$ and the \textit{queue length} $q\in\mathbb{R}^{4\times3}_{[0,1]}$, which are computed for each lane in the network. The former indicates the proportion of each lane's capacity currently occupied by vehicles, while the latter represents the normalized number of queued vehicles\footnote{A vehicle is said to be in queue if it's velocity $v$ falls below a specified threshold $\bar{v}$.} in each lane. Finally, the last state component takes into account the presence of pedestrians, and it is simply computed by counting the number of pedestrians $n\in\mathbb{N}$ currently waiting to cross the intersection.

We remark that all the described state components can be collected without the need for advanced sensors. Thanks to modern technologies, they can be readily available in most real-world scenarios through the use of advanced camera systems, without requiring invasive sensors, as cameras provide a non-intrusive and cost-effective solution for continuous monitoring of the traffic network.

Regarding the action space, we decided to extend the usual approach considered in the literature, i.e. maintaining the current phase or switching to the next one, in order to provide the RL agent with more flexibility. Specifically, we allow the agent to choose at each decision time the specific phase $a\in\mathcal{A}$ to activate, thus allowing the model to adapt to a wide variety of different scenarios.

\subsection{Reward function design}

To balance the needs of both vehicular and pedestrian traffic, we have designed a composite reward function, that accounts for the waiting time of both categories, as well as the stability of the traffic light phases:

\begin{equation}
    R = (1 - \beta) \cdot R_{veh} + \beta \cdot R_{ped} + R_{stab}
\end{equation}

Here, $R_{veh}=-(\sum_{v\in\mathcal{V}}t_v)/|\mathcal{V}|$ assigns a penalty based on the cumulative waiting time of vehicles. This penalty is normalized by the total number of vehicles currently in the environment ($ \mathcal{V}$), with $t_v$ indicating the travel time for vehicle $v$.
Analogously, $R_{ped}=-(\sum_{n\in\mathcal{N}}t_n)/|\mathcal{N}|$ applies a similar penalty for pedestrians. In this context, $\mathcal{N}$ denotes the set of pedestrians currently in the environment, and $t_n$ the waiting time of pedestrian $n$.

In this work, fairness is defined as the balanced allocation of service quality between different categories of road users, namely vehicles and pedestrians. Rather than exclusively optimizing vehicular throughput, the proposed reward formulation explicitly accounts for the waiting times experienced by both groups.
The parameter $\beta$ therefore acts as a fairness trade-off coefficient: it regulates the relative importance assigned to pedestrian and vehicular delays. Values of $\beta$ close to 0 prioritize vehicular efficiency, whereas values close to 1 prioritize pedestrian service. Intermediate values encourage more balanced solutions, preventing one category of users from being systematically disadvantaged.

From a multi-objective optimization perspective, $\beta$ can be interpreted as a scalarization parameter controlling the operating point along the Pareto frontier between vehicle and pedestrian waiting times.

Finally, the term $R_{stab}$ is needed to stabilize the traffic light phases, by introducing a penalty to discourage excessively frequent phase changes, promoting smoother and more stable traffic signal transitions:
\begin{align}
    R_{stab}=
    \begin{cases}
        -k \quad &\text{if } \mathcal{A}\subseteq\mathcal{A}_\tau  \\
        0 \quad &\text{otherwise.}
    \end{cases}
\end{align}
Here, $\mathcal{A}_\tau$ denotes the set containing the last $\tau$ actions performed by the RL agent, and $k$ is the assigned penalty.

As a final remark, we note that while implementing such a reward function demands additional data beyond typical state space features—posing challenges in real-world applications—it remains feasible within simulated environments. Here, we can readily access all necessary parameters to compute the reward. Consequently, during simulation-based training, this detailed information aids in optimizing algorithm performance without concern for its computation post-deployment in real-world settings, where such information is no longer required.

\begin{figure*}[t]
    \centering
    \begin{subfigure}[b]{0.49\textwidth}  
        \centering
        \includegraphics[width=\textwidth]{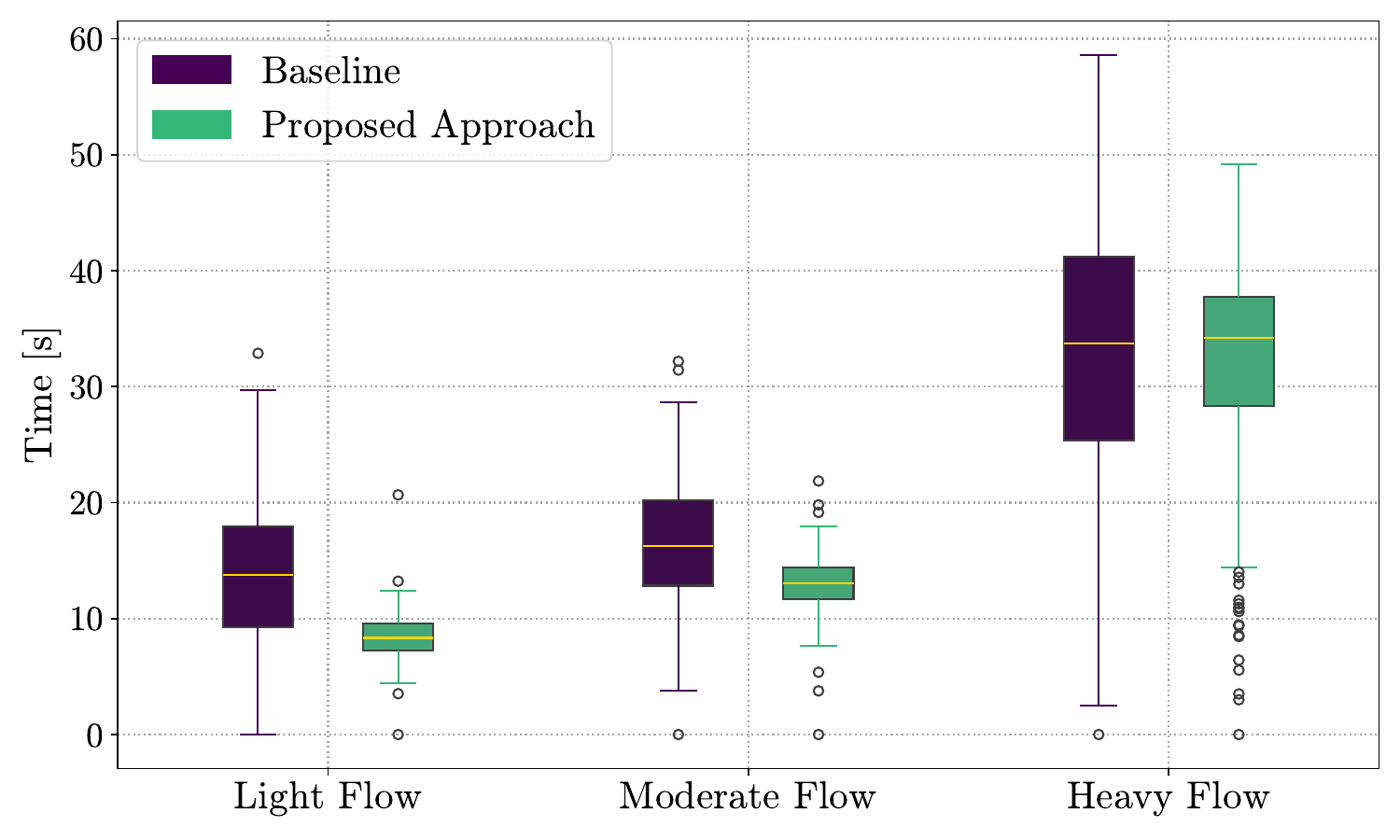}
        \caption{Vehicle waiting time}
        \label{fig:vehbaseline}
    \end{subfigure}
    \hfill
    \begin{subfigure}[b]{0.49\textwidth}  
        \centering
        \includegraphics[width=\textwidth]{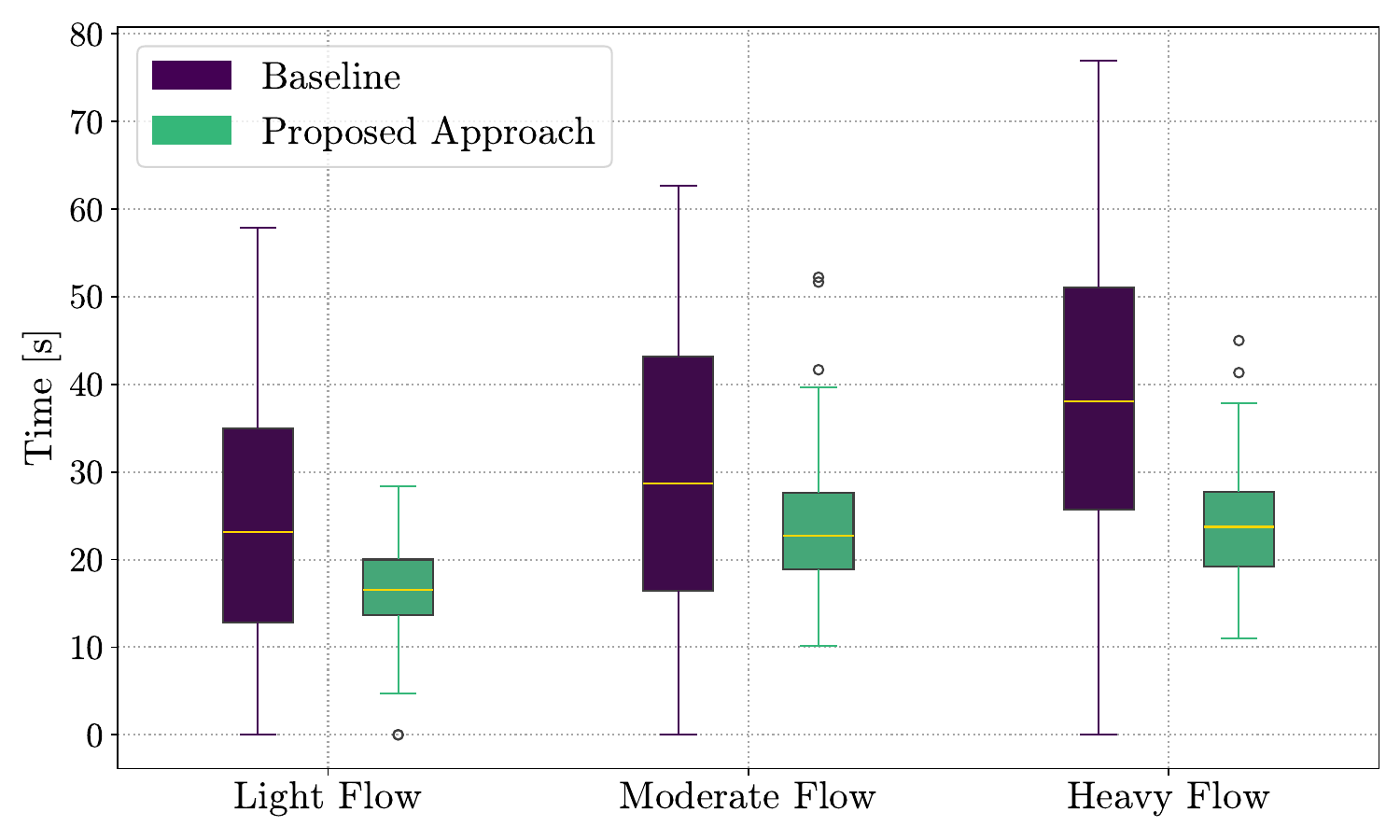}
        \caption{Pedestrian waiting time}
        \label{fig:pedbaseline}
    \end{subfigure}
    
    \caption{Comparison across different levels of traffic of vehicles (a) and pedestrians' (b) waiting times.}
    \label{fig:box1}
\end{figure*}

\begin{table}[H]
\begin{center}
\caption{Vehicles and pedestrians' flow rates}\label{tb:flow}
\begin{tabular}{ll}
\toprule
Light flow N/S & 750 veh/hour \\
Light flow E/W & 400 veh/hour \\
Moderate flow N/S & 850 veh/hour \\
Moderate flow E/W & 500 veh/hour \\
Heavy flow N/S & 1000 veh/hour \\
Heavy flow E/W & 600 veh/hour \\
Pedestrian flow N/S & 500 ped/hour \\
Pedestrian flow E/W & 300 ped/hour \\
\bottomrule
\end{tabular}
\end{center}
\end{table}

\section{Experiments}
\label{sec:exp}

To effectively train and evaluate our DDQN agent, choosing an appropriate simulation environment is crucial. For this project, we selected the FLOW platform (\cite{wu2021flow}), specifically tailored for reinforcement learning experiments in traffic lights control settings. FLOW integrates seamlessly with SUMO (Simulation of Urban MObility, \cite{SUMO2018}), a widely-used microscopic traffic simulator available under an open-source license. In our configuration, FLOW acts as the intermediary between SUMO and the RL framework by conforming to the standard Gymnasium APIs (\cite{towers_gymnasium_2023}), which are extensively adopted within the reinforcement learning community.


As previously introduced in Section \ref{sec:method}, our experiments were conducted on a large-scale four-way three-lane intersection scenario. The simulation operates with an internal time step of 1 second, while the RL agent's control step is fixed at 10 seconds. To fully leverage the strengths of the reinforcement learning paradigm, we trained our algorithm under various traffic conditions, mimicking realistic environments. Additionally, we assumed that the north-south route (N/S) experienced consistently higher traffic volumes than the east-west route (E/W), reflecting a common real-world situation where the main road intersects with a side road.

In terms of traffic light phases, we utilized the Barnes Dance scheme (\cite{chen2014relative}), enabling the agent to select from three primary phases: two dedicated to vehicles and one for pedestrians. Each main phase is followed by a yellow transition phase, which is not managed by the RL agent, as usually done in literature. A summary of the flow rates and simulation hyperparameters employed in this study are reported in Table \ref{tb:flow} and \ref{tb:params}, respectively\footnote{Regarding the DDQN hyperparameters, we employed standard values from the literature, which are not reported here for brevity.}.

\begin{table}[h]
\begin{center}
\caption{Simulation hyperparameters}\label{tb:params}
\begin{tabular}{ll}
\toprule
Action space dimension $|\mathcal{A}|$ & 3 \\
Training steps & $2e6$ \\
Flow rate change frequency (steps) & $2e3$ \\
$R_{stab}$ hyperparameters $(k, \tau)$ & (20, 3) \\
\bottomrule
\end{tabular}
\end{center}
\end{table}



    

\subsection{Baseline definition}

To evaluate our method's efficacy, we have compared it against a baseline consisting of a fixed-time traffic light. The cycle length for our baseline was determined using Webster's formula (\cite{webster1958traffic}), a well-established approach for calculating the optimal fixed-time cycle length of an intersection controller based on observed traffic volume. Utilizing the vehicle flow data from Table \ref{tb:flow}, we report the corresponding cycle lengths in Table \ref{tb:ph} below.

We selected a fixed-time traffic light controller as baseline because it represents a widely adopted solution in real-world urban deployments and provides an interpretable benchmark for evaluating adaptive traffic management strategies. While fairness-aware reinforcement learning approaches for traffic signal control have recently emerged in the literature, most existing methods focus exclusively on vehicular traffic and do not explicitly incorporate pedestrian flows. Extending and adapting these approaches to the mixed pedestrian-vehicle setting considered in this work would require substantial modifications to their state representations and reward formulations, and is therefore left as future work.

\begin{table}[h]
\caption{Optimal fixed-time traffic light phases as a function of the flow rate.}
\label{tb:ph}
\scriptsize
\begin{tabular}{ccc|ccc}
\toprule
{N/S} & {E/W} & {Pedestr.} & Light [s]  & Moderate [s] & Heavy [s] \\
\midrule
Green & Red & Red & 35 & 40 & 45 \\
Yellow & Red & Red & 5 & 5 & 5 \\
Red & Green & Red & 20 & 25 & 30 \\
Red & Yellow & Red & 5 & 5 & 5\\
Red & Red & Green & 15 & 15 & 15 \\
Red & Red & Yellow & 5 & 5 & 5\\
\bottomrule
\end{tabular}
\end{table}

\setcounter{figure}{3}
\begin{figure*}[t]
    \centering
    \begin{subfigure}[b]{0.49\textwidth}  
        \centering
        \includegraphics[width=\textwidth]{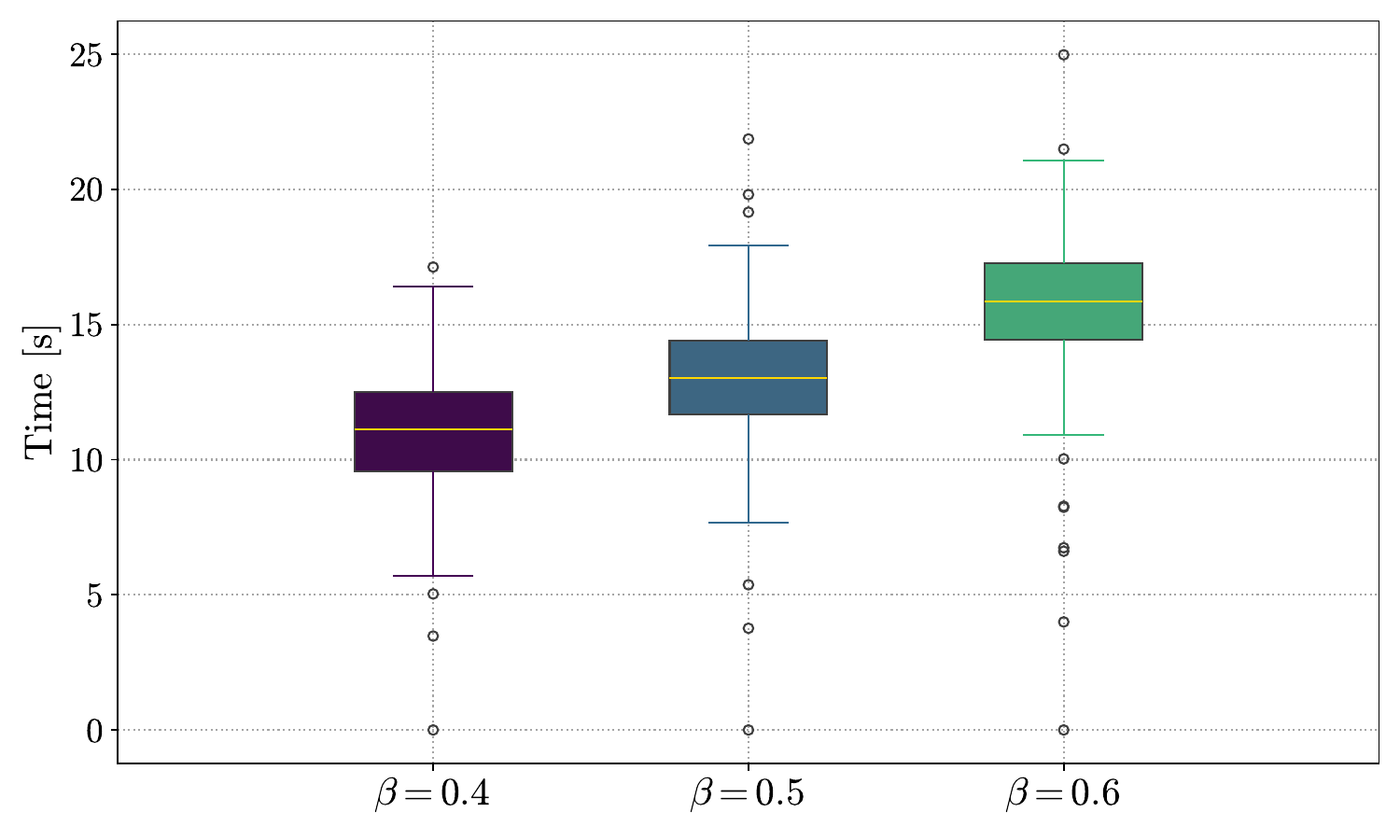}
        \caption{Vehicle waiting time}
        \label{fig:vehfair}
    \end{subfigure}
    \hfill
    \begin{subfigure}[b]{0.49\textwidth}  
        \centering
        \includegraphics[width=\textwidth]{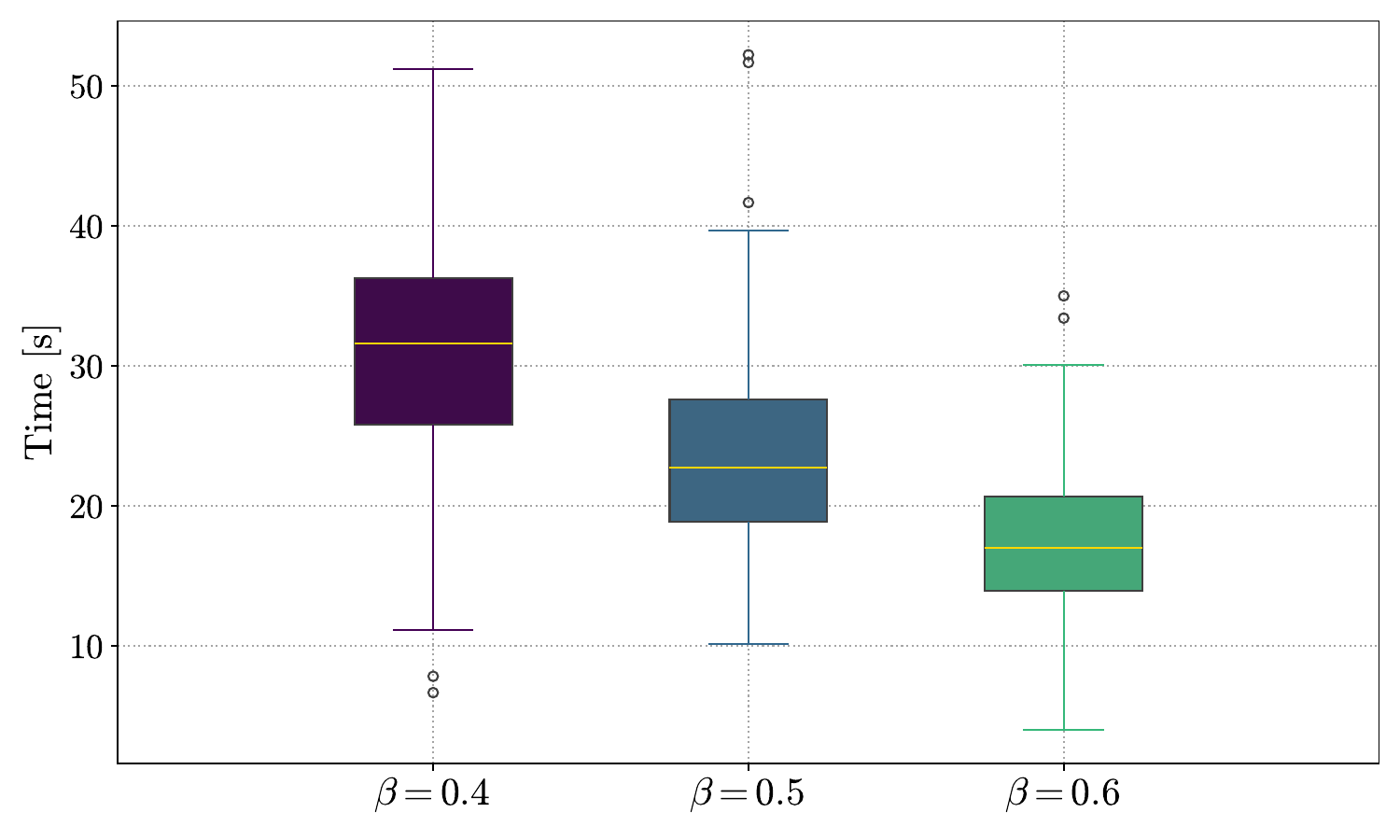}
        \caption{Pedestrian waiting time}
        \label{fig:pedfair}
    \end{subfigure}
    
    \caption{Comparison across different values of $\beta$ of vehicles (a) and pedestrians' (b) waiting times.}
    \label{fig:box2}
\end{figure*}

\setcounter{figure}{2}
\begin{figure}[H]
\begin{center}
\includegraphics[width=\linewidth]{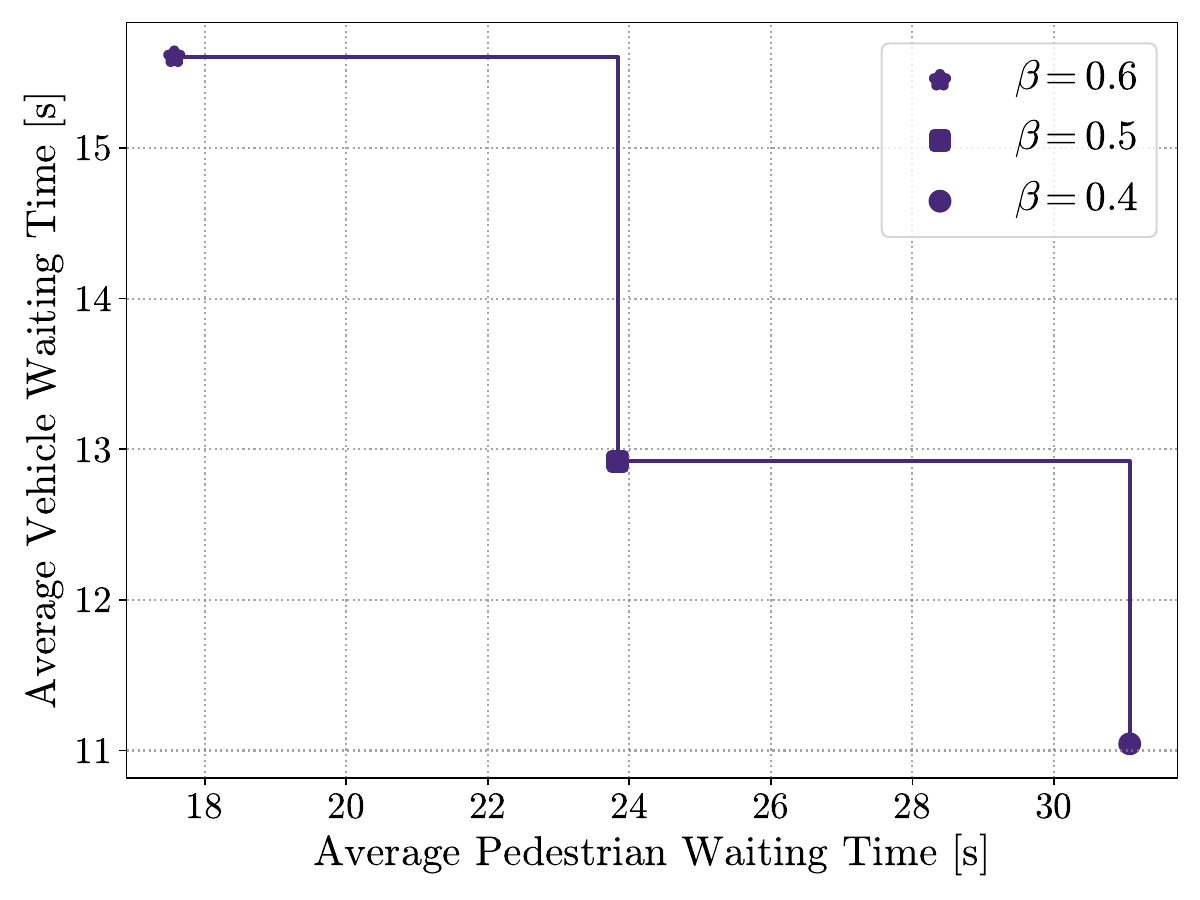}    
\caption{Pareto frontier for the considered multi-objective RL problem. The average pedestrian and vehicle waiting times are represented on the x and y axes, respectively. Each point on the front corresponds to a different value of $\beta$.} 
\label{fig:pareto}
\end{center}
\end{figure}

\subsection{Simulation results}

We assess the performance of our algorithm in comparison to the baseline introduced above by conducting an evaluation phase lasting $20000$ time steps (approximately 5.5 hours of simulation time). During this phase, the incoming vehicle flow rate is periodically varied according to the three configurations presented in Table~\ref{tb:flow}. As shown in Figure~\ref{fig:box1}, our method consistently reduces the individual waiting times for both vehicles and pedestrians across all traffic configurations, outperforming the optimal fixed-time traffic light baseline. Importantly, the advantage of our approach extends beyond the observed reduction in waiting times: indeed it leverages a single reinforcement learning agent that \textit{autonomously} adapts to varying traffic conditions, without requiring manual adjustments. In contrast, the baseline requires explicit reconfiguration of the traffic light cycle length each time the traffic flow varies, an impractical requirement in real-world deployments.

Moving further, we now present the discussion related to a crucial contribution of our work, i.e. ensuring fairness in the prioritization of vehicles and pedestrians. While for the previous experiments we fixed the fairness temperature $\beta=0.5$, for the following analysis we have trained three different reinforcement learning agents, varying the value of $\beta$ across the set $\mathrm{B}=\{0.4, 0.5, 0.6\}$, and testing their performances in the same evaluation setup described before. As we can see from Figure \ref{fig:pareto}, all the three agents achieve Pareto optimal performances, considering as minimizing objectives the average individual waiting times of vehicles and pedestrians. As expected, higher values of $\beta$ tend to prioritize more pedestrians, and the opposite happens if $\beta$ decreases. Furthermore, to offer deeper insights into the robustness of the obtained Pareto efficient solutions, we reported in Figure \ref{fig:box2} the distributions of both the vehicles and pedestrians' waiting times for the three obtained solutions. As expected, they have opposite trends as $\beta$ increases, consistently with the above discussion. We remark that we could easily extend our analysis by considering a larger set of Pareto-efficient solutions, but we decided instead to focus on a reduced, representative subset since, in practical scenarios, decision-making benefits more from a few high-quality options— around key trade-off regions—than from the full front.

In conclusion, these results confirm the effectiveness and adaptability of using different models tailored to various scenarios: situations that demand greater consideration and attention for vulnerable road users can benefit from models with higher beta values; conversely, scenarios where road traffic reaches critical and challenging levels can leverage models with lower beta values to reduce traffic congestion. Most importantly, these models can be selected online by the service provider, depending on the situation.

\section{Conclusions and future directions}
\label{sec:conc}

In this work, we proposed a reinforcement learning-based traffic light control agent that incorporates fairness-aware considerations between vehicles and pedestrians into its decision-making process. Unlike traditional approaches that focus solely on traffic efficiency, our method explicitly accounts for the needs of different categories of road users, aiming to avoid the systematic over-prioritization of one group over the other. By integrating both vehicular and pedestrian waiting times into the reward formulation, the proposed approach promotes a more balanced allocation of green time, while dynamically adapting to varying traffic conditions.

We validated our approach through extensive simulations, demonstrating its ability to outperform standard fixed-time traffic light controllers across different traffic configurations, while promoting more equitable service trade-offs between vehicles and pedestrians. Importantly, our agent achieves this using realistic observations, supporting the potential applicability of the proposed framework in real-world urban environments.

We believe that this study represents a step forward toward fairness-aware urban mobility management. In future work, we plan to evaluate the proposed method on real-world traffic data and extend the framework by integrating environmental sustainability objectives, such as the reduction of CO$_2$ emissions, while continuing to balance the service needs of both vehicles and pedestrians.


\bibliography{ifacconf}


\end{document}